\documentclass{ieeeaccess}

\usepackage{cite}
\usepackage{amsmath,amssymb,amsfonts}
\usepackage{graphicx}
\usepackage{textcomp}

\usepackage{times,epsfig}
\usepackage{amsthm, url}
\usepackage{latexsym}
\usepackage{lipsum}
\usepackage{balance}
\usepackage{subfigure}
\usepackage{nicefrac}
\usepackage{booktabs}
\usepackage{color}
\usepackage[acronym,nohypertypes={acronym}]{glossaries}
\usepackage[stable]{footmisc}
\usepackage[table,xcdraw]{xcolor}
\usepackage[utf8]{inputenc}
\usepackage[colorlinks,backref]{hyperref} 
\usepackage[inline]{enumitem} 
\usepackage[export]{adjustbox}
\usepackage[numbers]{natbib}
\usepackage{numprint}

\usepackage{caption}

\def\BibTeX{{\rm B\kern-.05em{\sc i\kern-.025em b}\kern-.08em
    T\kern-.1667em\lower.7ex\hbox{E}\kern-.125emX}}

\newacronym{osnn}{OSNN}{Open-Set Nearest Neighbors}
\newacronym{ssvm}{SSVM}{Specialized Support Vector Machines}
\newacronym{pisvm}{PISVM}{\glstext{svm} with Probability of Inclusion}
\newacronym{wsvm}{WSVM}{Weibull-calibrated \glstext{svm}}
\newacronym{svm}{SVM}{Support Vector Machines}
\newacronym{ocsvm}{OCSVM}{One-class \glstext{svm}}
\newacronym{dbc}{DBC}{Decision Boundary Carving}
\newacronym{ncm}{NCM}{Nearest Class Mean with cosine distance}
\newacronym{et}{ET}{extremely randomized trees, a.k.a., Extra-Trees}
\newacronym{psvm}{PSVM}{\glstext{svm} with Platt's probability for rejection}
\newacronym{2psvm}{2PSVM}{2-phase \glstext{svm}}
\newacronym{softmax}{SOFTMAX}{thresholding softmax probability}

\newacronym[first={\glsentrydesc{dresden}}]{dresden}{Dresden}{Dresden Image Database}
\newacronym{isa}{ISA Unicamp}{Image Source Attribution Unicamp}
\newacronym[first={\glsentrydesc{flickr}}]{flickr}{Flickr}{Flickr Unicamp}

\newacronym{na}{NA}{Normalized Accuracy}
\newacronym{aks}{AKS}{Accuracy on Known Samples}
\newacronym{aus}{AUS}{Accuracy on Unknown Samples}
\newacronym{osfmM}{OSFM$_{\mathrm{M}}$}{open-set macro-averaging f-measure}
\newacronym{osfmm}{OSFM$_{\mu}$}{open-set micro-averaging f-measure}
\newacronym{fmM}{FM$_{\mathrm{M}}$}{traditional binary-based macro-averaging f-measure}
\newacronym{fmm}{FM$_{\mu}$}{traditional binary-based micro-averaging f-measure}
\newacronym{da}{DA}{Detection Accuracy}
\newacronym{dks}{DKS}{Detection on Known Samples}
\newacronym{dus}{DUS}{Detection on Unknown Samples}
\newacronym[symbol={\mathrm{\glsentrytext{tp}}}]{tp}{TP}{true positive}
\newacronym[symbol={\mathrm{\glsentrytext{fp}}}]{fp}{FP}{false positive}
\newacronym[symbol={\mathrm{\glsentrytext{fn}}}]{fn}{FN}{false negative}

\newacronym{networkopen}{\mbox{NetOpen}}{Network Open}
\newacronym[first={\glsentrytext{open}}]{open}{\mbox{Open}}{??}
\newacronym[first={\glsentrytext{closed}}]{closed}{\mbox{Closed}}{??}

\newacronym{cnn}{CNN}{Convolutional Neural Network}

\newacronym{klos}{KLOS}{known-labeled open space}

\newacronym[first={\emph{\glsentrytext{ku}}~\cite{Bendale2015}}]{ku}{known unknown}{??}

\def\x{{\mathbf x}}
\def\I{{\mathbf I}}
\def\f{{\mathbf{f}}}
\def\C{{\mathcal{C}}}
\def\Ck{{\C_\text{known}}}
\def\co{{c_0}}                  
\def\M{{\mathcal{M}}}

\def\fconv{\textnormal{\ensuremath{{\mathbf{f}_\text{conv} }}}}
\def\fipone{\textnormal{\ensuremath{{\mathbf{f}_\text{ip1} }}}}
\def\fiptwo{\textnormal{\ensuremath{\mathbf{f}_\text{ip2} }}}
\def\frich{\textnormal{\ensuremath{{\mathbf{f}_\text{rich} }}}}
\def\fcfa{\textnormal{\ensuremath{{\mathbf{f}_\text{cfa} }}}}
\newacronym[first={\glsentrytext{conv}}]{conv}{\fconv{}}{??}
\newacronym[first={\glsentrytext{ip1}}]{ip1}{\fipone{}}{??}
\newacronym[first={\glsentrytext{ip2}}]{ip2}{\fiptwo{}}{??}
\newacronym[first={\glsentrytext{rich}}]{rich}{\frich{}}{??}
\newacronym[first={\glsentrytext{cfa}}]{cfa}{\fcfa{}}{??}

\def\striplastbar#1{\striplastbara{#1}#1\end /\end\eend}
\def\striplastbara#1#2/\end#3\eend{\ifx\end#3\end#1\else#2\fi}
\newcommand{\changedir}[1]{\striplastbar{#1}__}

\npdecimalsign{.}
\npfourdigitnosep{}
\newcommand\num[1]{#1}          
\renewcommand\num[1]{\nprounddigits{4}\numprint{#1}}          

\newcommand\mynum[1]{\num{#1}}
\newcommand\mynummax[1]{\textbf{\num{#1}}} 
\newcommand\mynummin[1]{\textit{\num{#1}}} 

\newenvironment{enumerateromaninline}
{\begin{enumerate*}[label={(\roman*)}]}
  {\end{enumerate*}}

\begin{document}

\history{Date of publication June 6, 2019, date of current version June 4, 2019.}
\doi{10.1109/ACCESS.2019.2921436}
\def\thevol{0}
\def\theyear{2019}

\title{An In-Depth Study on Open-Set\\Camera Model Identification}

\author{\uppercase{Pedro~Ribeiro~Mendes~J\'{u}nior}\authorrefmark{1},
		\uppercase{Luca~Bondi}\authorrefmark{2},~\IEEEmembership{Student~Member,~IEEE},
        \uppercase{Paolo~Bestagini}\authorrefmark{2},~\IEEEmembership{Member,~IEEE},
        \uppercase{Stefano~Tubaro}\authorrefmark{2},~\IEEEmembership{Senior~Member,~IEEE},
        and~\uppercase{Anderson~Rocha}\authorrefmark{1},~\IEEEmembership{Senior~Member,~IEEE}}
\address[1]{Institute of Computing, University of Campinas (Unicamp), Av. Albert Einstein, 1251, CEP 13083-852, Campinas, S\~ao Paulo, Brazil (e-mail: pedrormjunior@gmail.com / anderson.rocha@ic.unicamp.br).}
\address[2]{Dipartimento di Elettronica, Informazione e Bioingegneria, Politecnico di Milano, Piazza Leonardo da Vinci 32, 20133, Milan, Italy (e-mail: luca.bondi / paolo.bestagini / stefano.tubaro@polimi.it).}
\tfootnote{%
  This material is based on research sponsored by DARPA and Air Force Research Laboratory (AFRL) under agreement number FA8750-16-2-0173.
  The U.S. Government is authorized to reproduce and distribute reprints for Governmental purposes notwithstanding any copyright notation thereon.
  The views and conclusions contained herein are those of the authors and should not be interpreted as necessarily representing the official policies or endorsements, either expressed or implied, of DARPA and Air Force Research Laboratory (AFRL) or the U.S. Government.
  This study was financed in part by the Coordenação de Aperfeiçoamento de Pessoal de Nível Superior---Brasil (CAPES)---Finance Code 001.
  This work was supported in part by S\~{a}o Paulo Research Foundation (FAPESP) under the grant \#2017/12646-3 (D\'{e}j\`{a}Vu project), and CAPES DeepEyes project.}

\markboth
{Mendes~J\'{u}nior \headeretal: An In-Depth Study on Open-Set Camera Model Identification}
{Mendes~J\'{u}nior \headeretal: An In-Depth Study on Open-Set Camera Model Identification}

\corresp{Corresponding author: Pedro~Ribeiro~Mendes~J\'{u}nior (e-mail: pedrormjunior@gmail.com).}

\begin{abstract}
Camera model identification refers to the problem of linking a picture to the camera model used to shoot it.
As this might be an enabling factor in different forensic applications to single out possible suspects (e.g., detecting the author of child abuse or terrorist propaganda material), many accurate camera model attribution methods have been developed in the literature.
One of their main drawbacks, however, is the typical closed-set assumption of the problem.
This means that an investigated photograph is always assigned to one camera model within a set of known ones present during investigation, i.e., training time.
The fact that a picture can come from a completely unrelated camera model during actual testing is usually ignored.
Under realistic conditions, it is not possible to assume that every picture under analysis belongs to one of the available camera models.
To deal with this issue, in this paper, we present an in-depth study on the possibility of solving the camera model identification problem in open-set scenarios.
Given a photograph, we aim at detecting whether it comes from one of the known camera models of interest or from an unknown one.
We compare different feature extraction algorithms and classifiers specially targeting open-set recognition.
We also evaluate possible open-set training protocols that can be applied along with any open-set classifier, observing that a simple alternative among the selected ones obtains the best results.
Thorough testing on independent datasets shows that it is possible to leverage a recently proposed convolutional neural network as feature extractor paired with a properly trained open-set classifier aiming at solving the open-set camera model attribution problem even on small-scale image patches, improving over state-of-the-art available solutions.
\end{abstract}
\begin{keywords}
Camera model identification, image forensics, open-set recognition, open-set training protocol.
\end{keywords}

\titlepgskip=-15pt

\maketitle

\section{Introduction}\label{sec:intro}
\PARstart{F}{rom} social networks to media sharing platforms, digital pictures are spreading all over the Internet at an overgrowing pace.
However, a major drawback of this phenomenon is the diffusion of illicit or illegal material online, specially visual content.
In order to fight this trend, multimedia forensic researchers have focused on the development of numerous solutions aiming at inferring pieces of information related to the acquisition and editing history of images~\cite{Stamm2013, Piva2013, Rocha2011}, among others.

A common problem of interest for forensic analysts is camera model identification.
This means being capable of detecting which camera model has been used to shoot a given digital photograph based solely on its content.
Indeed, this is a first step toward tracking down the author of distributed illicit contents \cite{Kirchner2015} (e.g., pictures related to acts of violence, images linked to terrorist behavior, sexually exploitative imagery of children, among others).
Given the social relevance of this problem, in the last few years, a continuous effort has been put forward to the development of more accurate and efficient camera model identification solutions.
These can be broadly split into two categories:
(i) model-based methods leveraging the study of characteristic traces left behind by specific operations applied by different camera models on acquired images;
and (ii) data-driven methods based on machine-learning techniques that seek to ``learn'' the patterns of such telltales automatically.
Considering the first category, we can cite methods relying on traces left by color filter array (CFA) interpolation~\cite{Bayram2005, Cao2010, Zhao2016}, on histogram equalization footprints~\cite{Chen2007a}, on traces left by camera lenses~\cite{Choi2006}, and on characteristic noise analysis~\cite{Thai2014}.
Considering the second category, in turn, we can cite the works of \citet{Chen2015a, Marra2015}, and \citet{Tuama2016}, which extract statistical features in the pixel-domain to train supervised machine-learning classifiers specialized at the problem.
More recently, relying upon  advancements on deep learning techniques, data-driven solutions based on \glspl{cnn} have outperformed prior art \cite{Tuama2016a, Bondi2017, Bondi2017a, Ferreira2018a, Bayar2017a}, and are becoming an area's staple.

\begin{figure*}[t]
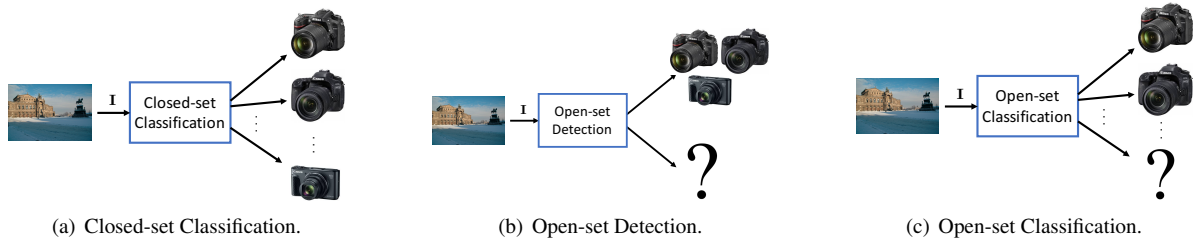

	\centering
	\subfigure[Closed-set Classification.]{\includegraphics[width=.25\linewidth]{\changedir{figures/}closed_classification}\label{fig:closed_classification}} \hfil
	\subfigure[Open-set Detection.]{\includegraphics[width=.25\linewidth]{\changedir{figures/}open_detection}\label{fig:open_detection}} \hfil
	\subfigure[Open-set Classification.]{\includegraphics[width=.25\linewidth]{\changedir{figures/}open_classification}\label{fig:open_classification}}
	\caption{Different camera model identification problem formulations.  \subref{fig:closed_classification}~In the closed-set problem, an image is attributed to a known camera model.  \subref{fig:open_detection}~In the open-set detection problem, an image is detected as belonging either to the known set or the unknown set of models.  \subref{fig:open_classification}~In the open-set classification problem faced in this work, an image is detected as belonging to the known or unknown model sets and, in the first case, the correct model is also estimated.}
	\label{fig:problem}
\end{figure*}

The drawback of all aforementioned data-driven techniques---or, more precisely, the evaluation setup to validate such techniques---is that they mainly cope with camera model identification in a closed-set setup.
This means that a finite set of camera models is considered when designing the solution, and each image is attributed to one of these models.
However, oftentimes analysts must work in open-set scenarios.
This means that the investigator must also be able to recognize whether an image does not belong to any of the known models of interest \cite{Kirchner2015}.

In this vein, we present herein an in-depth study on open-set camera model attribution based on a supervised learning pipeline.
Specifically, we focus on methodologies that perform an analysis at patch level rather than on the whole image, as this opens the door to future development of tampering detection and localization methods as shown by \citet{Bondi2017b}.
To the best of our knowledge, open-set camera model attribution has only been introduced by \citet{Gloe2012a} and later on approached by \citet{Bayar2018}.
\citet{Bayar2018} focus on an open-set binary detection problem, i.e., detecting whether an image comes from a known or unknown camera model.
Conversely, we aim to solve the joint problem of (i) detecting whether the image under analysis comes from a known or from an unknown camera model and (ii) determining the image source model when it comes from the set of known models.

In previous work~\cite{Jain2014}, a general open-set classifier have been proposed along with \emph{cross-class validation}, which is a method tailored to open-set scenarios that aims at searching for the parameters of the proposed open-set classifier.
In parallel, another previous work~\cite{MendesJunior2017}, also proposing an open-set classifier, introduced a \emph{parameter optimization} procedure that is also tailored at searching the parameters of their proposed classifier, which shares the same essence of the cross-class validation.
In the latter work, authors have suggested as future work the employment of their parameter optimization method as a general grid-search procedure that could be applied to any open-set classifier.
In our work, we follow this direction and we evaluate what we call \gls{closed} training protocol (the traditional form) and the \gls{open} training protocol (with the same essence of the cross-class validation~\cite{Jain2014} and the parameter optimization~\cite{MendesJunior2017}).
We further study those alternatives and we formalize and evaluate what we call the \gls{networkopen} training protocol, specifically tailored to situations in which deep features are employed.
As we shall see later on along with the presented results, the equivalence of \gls{open} and \gls{networkopen} indicates the \gls{open} training protocol as the best and cheaper alternative in terms of data required to be employed.

In light of these considerations, our key contributions are the following:
\begin{itemize}
	\item We study the open-set camera model identification problem analyzing state-of-the-art open-set classification methods.
	\item We evaluate the effectiveness of \glspl{cnn} features, compared to hand-crafted ones, for per-patch classification in open-set setups.
	\item We formalize and evaluate open-set training protocols applied to open-set classification methods during training for proper estimate of parameters for the open-set scenario.
	\item We carry out the first large-scale testing on the open-set camera model identification problem considering independent datasets and several algorithms, also comparing with known solutions in the literature~\cite{Bayar2018}.
\end{itemize}
The best evaluated solution for the problem combines a deep feature extraction method and a state-of-the-art open-set classifier trained with an open-set training protocol of intermediate complexity.
This solution works on $64\times64$ color patches, making it useful for forgery localization techniques~\cite{Bondi2017b}.
Moreover, it is capable of reaching state-of-the-art accuracy also in the closed-set framework.

The rest of the paper is structured as it follows.
Section~\ref{sec:problem} formally introduces the camera model identification problem under different points of view.
Section~\ref{sec:algorithm} provides all the details about the algorithmic pipeline used in our evaluation.
Section~\ref{sec:setup} reports information about the considered experimental setup.
Section~\ref{sec:results} presents the performed experiments and achieved results.
Finally, Section~\ref{sec:conclusions} concludes the paper.

\section{Open-set Camera Model Identification Problem}\label{sec:problem}
In this section, we introduce the problem of camera model identification, from the closed-set to the open-set one faced in this paper.

Camera model identification generally refers to the problem of assigning an image, in a blind fashion, to the camera model that was used to shoot it.
This means that no watermarks or side information such as header or EXIF data are used, assuming they will not be available during investigation.
Depending on the considered constraints, camera model identification can be cast into different kinds of problems, as shown in Figure~\ref{fig:problem}.
In the following, we report the main differences between these problem formulations.

\subsection{Closed-set Classification}
Closed-set camera model classification is the problem of assigning an image to a camera model within a known set of possible models, as depicted in Figure~\ref{fig:closed_classification}.
In this scenario, it is required to assume that the investigator is sure that the camera model of the picture under analysis belongs to the set of candidate models.

Formally, let $\I$ be a color image acquired with the camera model identified by label $c$.
Consider further $\Ck$ as the set of labels $c$ belonging to the known camera model dataset, e.g., available to the analyst when developing the solution.
The goal in closed-set camera model identification is to estimate the label $\hat{c} \in \Ck$ associated to the picture under analysis.

This is by far the most widely considered scenario in the literature~\cite{Kirchner2015}.
However, closed-set classification is bound to fail whenever the analyst has no full knowledge on all the possible used camera models: in real-case open-set scenarios, it happens that $c \in \Ck \cup \{\co\}$, in which $\co$, $\co \notin \Ck$, is the \emph{unknown label} that represents any unknown class.

\subsection{Open-set Detection}
Relaxing the constraint of knowing all possible camera models, we enter the open-set realm.
Indeed, in an open-set scenario, the image under analysis can belong to either known or unknown camera models.
In particular, we refer to open-set camera model detection as the problem of detecting whether an image belongs to the set of known models, or to the set of unknown ones, as depicted in Figure~\ref{fig:open_detection}.

Formally, the goal of open-set camera model detection is to estimate whether $c \in \Ck$ or $c \notin \Ck$ for a given image $\I$.
This is basically a two-class classification problem that does not provide the analyst with information on the actual used camera model.
To infer the possible used camera model, an open-set detection solution should be paired with a subsequent step of closed-set classification, as proposed by \citet{Bayar2018}.

\subsection{Open-set Classification}\label{sec:open-set-classification}
The most complete camera model identification problem formulation is that of open-set classification.
As a matter of fact, this refers to the problem of jointly estimating whether the image under analysis comes from a camera in the known set of models or from an unknown model and, if condition one holds, also detecting which model it is, as depicted in Figure~\ref{fig:open_classification}.

Formally, the goal of open-set camera model identification is to estimate $\hat{c} \in \Ck \cup \{\co\}$ for a given image $\I$.

Typically, to properly develop an open-set classification solution, three different kinds of data are employed:
\begin{itemize}
	\item \textbf{Known data (train and test):} images shot with models $c \in \Ck$ that the analyst must correctly detect and classify.
	\item \textbf{Known-unknown data (optional; train and test):} images shot with models available at training time but assumed as unknown in order to model unknown camera models at algorithm validation time.
          Those data might or might not be available.
	\item \textbf{Unknown-unknown data (test only):} images shot with models $c \notin \Ck$ and not used for either training or validation, used to properly evaluate a method's performance in the wild.
          Those data only appear for classification once the classifier is trained.
\end{itemize}

Open-set classification is by far the most complete problem formulation of the overall camera model identification problem.
In this paper, we present an algorithmic pipeline to solve this problem, deeply analyzing each building block of the algorithm in all combinations of the alternatives.

Previous works in open-set camera model identification have not fully evaluated the multiclass open-set classification problem.
\citet{Bayar2018} have considered the performance of the classification methods for detecting known vs unknown and, independently, the closed-set classification performance among the classes.
In this latter evaluation, the classifiers work in a closed-set scenario, i.e., they never predicts as unknown.
It is worth considering that the accuracy in a problem as described in Section~\ref{sec:open-set-classification} tends to be smaller than considering, independently, the detection accuracy and the closed-set accuracy of the methods without the option for rejection, as in the open-set classification problem the classification methods can perform the following types of error: \emph{misclassification}, \emph{false unknown}, and \emph{false known}~\cite{MendesJunior2017}.

\section{Evaluation Pipeline}\label{sec:algorithm}
In this section, we provide all the details about the factors we evaluate in this work.
We first provide an overview of the overall algorithmic pipeline.
Then, we focus on each separate block of it, reporting information about all methodologies employed in this paper.

\subsection{Pipeline}
\begin{figure}[t]
	\centering
	\includegraphics[width=0.9\columnwidth]{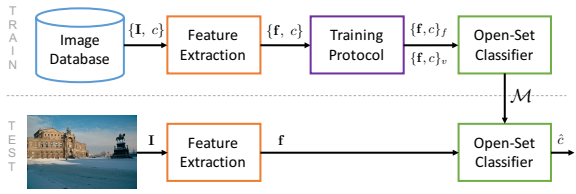}
	\caption{Supervised pipeline for open-set camera model identification. At training time (top), a classifier learns how to associate features $\f$ extracted from labeled images $\I$ to labels $c$. At testing time (bottom), the learned model $\M$ is used to estimate the label $\hat{c}$ from the image under analysis.}
	\label{fig:pipeline}
\end{figure}

To solve open-set camera model attribution, we study the possibility of exploiting a supervised classification strategy leveraging image descriptors tailored to capture camera-based traces proposed in the closed-set scenario literature.
Specifically, we follow the pipeline depicted in Figure~\ref{fig:pipeline}, which is composed by three main modules:
\begin{enumerateromaninline}
\item a \emph{feature extractor},
\item a \emph{training protocol} for preparing training data, and
\item an \emph{open-set classifier}.
\end{enumerateromaninline}
For each module, we investigate the possibility of using different strategies.

\begin{figure*}[t]
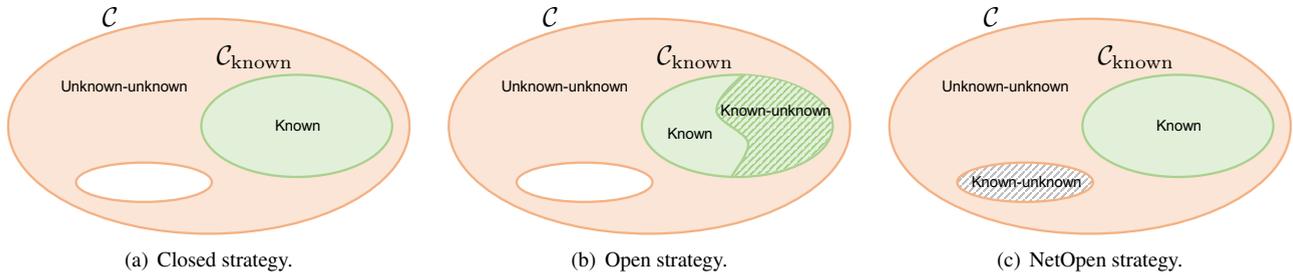

	\centering
	\subfigure[\gls{closed} strategy.]{\includegraphics[width=.3\linewidth]{\changedir{figures/}closed}\label{fig:closed}} \hfil
	\subfigure[\gls{open} strategy.]{\includegraphics[width=.3\linewidth]{\changedir{figures/}open}\label{fig:open}} \hfil
	\subfigure[\glstext{networkopen} strategy.]{\includegraphics[width=.3\linewidth]{\changedir{figures/}netopen}\label{fig:netopen}}
	\caption{Different training strategies: \subref{fig:closed}~\gls{closed}, \subref{fig:open}~\gls{open}, and \subref{fig:netopen}~\glstext{networkopen}.
          The set $\C$ contains all camera models, whereas $\Ck$ only contains known models.
          Green regions represent data for training the \gls{cnn} employed for feature extraction.
          Striped regions represent known-unknown camera models employed for classifiers' parameter search and also for training final model in case of~\subref{fig:open}.
          Orange regions represent unknown-unknown camera models that appear only for testing.
          Blank regions are never employed for testing for fair strategies comparison.
          Although not represented, test set also contains instances from known camera models, for proper evaluation of the classifiers.
        }
	\label{fig:strategies}
\end{figure*}

\emph{Feature extraction} consists in computing a discriminative feature vector $\f$ from an image $\I$.
The feature extractor algorithm is tuned to obtain characteristic camera model information while compacting data dimensionality. Feature vectors $\f$ extracted from pictures sharing the same camera model should be similar.
Conversely, feature vectors $\f$ extracted from images shot with different models should be, ideally, strongly dissimilar.

Open-set classifiers, as we shall see, tend to associate a \emph{bounded region} of the feature space to the known classes.
A recent work~\cite{MendesJunior2017} has shown that the split of training data for parameter search can have an influence on the final model obtained by an open-set classifier.
The \emph{training protocol} splits the training data $\{\f, c\}$ into fitting data $\{\f, c\}_{f}$ and validation data $\{\f, c\}_{v}$ for parameters search, as depicted in Figure~\ref{fig:pipeline}.
This is a delicate step, as a good open-set classifier must ``learn'' its parameters taking into account the \emph{risk of the unknown}, not just the \emph{empirical risk} measured on known data~\cite{Scheirer2013}.
In essence, prominent alternatives at this stage aim at employing part of the known training data as known-unknown data, as a form of simulation of the unknown.

The role of an \emph{open-set classifier} is to learn a mapping between feature vectors $\f$ and camera labels $c$.
This mapping is learned at training time by observing several different pairs $\{\f, c\}$ for many different images and $c$ values, $c \in \Ck$.
The open-set classifier partitions the space spanned by all possible vectors $\f$, associating different regions of the feature space to different labels $c \in \Ck \cup \{\co\}$.

Once the system has been fully trained, it can be deployed.
Whenever a new image $\I$ under investigation is considered, a feature vector $\f$ is extracted.
The open-set classifier model $\M$ is employed to predict the vector $\f$ with one class label $\hat{c} \in \Ck \cup \{\co\}$.

\subsection{Feature Extractors}
Different feature extractors for camera-related features have been proposed in the literature.
We decided to focus on recently proposed ones that have shown good performance in closed-set camera model attribution setups.

\subsubsection{Rich features}
\citet{Fridrich2012} have proposed the use of statistical descriptors known as rich features for steganalysis.
Rich features are obtained by preprocessing an image through high-pass filtering, quantization and truncation.
The rich feature vector is then computed by counting the occurrences of different pixel group combinations.
The use of rich features has subsequently proved successful for other forensic applications, from tampering detection \cite{Cozzolino2014} to camera model attribution \cite{Marra2015}.
We denote $\frich \in \mathbb{R}^{338}$ as the rich feature vector referred to as SPAM by \citet{Marra2015} for camera model identification.
It has already proved to be more discriminative than those proposed by \citet{Gloe2012a, Xu2012}, and \citet{Celiktutan2008} as shown by \citet{Marra2015}.

\subsubsection{CFA features}
As shown by \citet{Chen2015a}, the concept of rich features can be extended to work across different image color planes.
\citet{Chen2015a} have shown that it is possible to capture characteristics related to color filter arrays (CFA) for camera model identification.
For this reason, we denote $\fcfa \in \mathbb{R}^{1372}$ as the CFA-based feature vector proposed by~\citet{Chen2015a}.
As shown by~\citet{Bondi2017}, this can be considered a baseline solution especially when large images are concerned.

\subsubsection{\texorpdfstring{\gls{cnn}}{CNN}-derived features}
We adopt as a data-driven method the \gls{cnn} proposed by \citet{Bondi2017} with an architecture comprising four convolutional layers followed by two inner product layers.
It has been successfully applied to attribute images to 18 different camera models using $64 \times 64$ patches as input.
In principle, the output of each \gls{cnn} layer can be employed as a feature vector $\f$.
We employ three layers in this work:
\begin{enumerateromaninline}
\item $\fconv \in \mathbb{R}^{128}$, obtained after the last convolutional layer;
\item $\fipone \in \mathbb{R}^{128}$, obtained after the first inner product layer; and
\item $\fiptwo \in \mathbb{R}^{n}$, obtained after the second inner product layer, where $n = |\Ck|$ is the cardinality of the set of known cameras (18 in our experiments, as in the work of \citet{Bondi2017}).
\end{enumerateromaninline}

\subsection{Training Protocols}\label{sec:training-protocol}
To train open-set classifiers, a set of hyper-parameters must be tuned through some method of parameter search to maximize classification accuracy and generalization/specialization capabilities of the employed method.
A typical way to do this consists in splitting training data $\{\f, c\}$ into \emph{fitting} $\{\f, c\}_{f}$ and \emph{validation} $\{\f, c\}_{v}$ data.
The selected classifier is then trained on fitting data using different sets of hyper-parameters.
Finally, the parameters which model provides the highest accuracy on the set of validation data are selected.
The final model is generated on the entire training set with those parameters and results are reported on images belonging to a completely separate (independent) test dataset.
In this work, we explore three different training strategies for open-set classifiers.
The introduction of this stage in the pipeline was inspired by the work of~\citet{MendesJunior2017}, which pointed out their \emph{parameter optimization} as a general form of grid search for future investigation.
In Figure~\ref{fig:strategies}, we depict those alternatives as described below.

\subsubsection{\texorpdfstring{\gls{closed}}{Closed} strategy}
Depicted in Figure~\ref{fig:closed}, this is the simplest training strategy, in which no knowledge on the unknown classes is simulated.
Indeed, both fitting and validation datasets contain samples from all $n$ known classes (i.e., camera models), and no instance from known-unknown data is used in validation.
In other words, parameter search is performed simulating a closed-set setup.
This means that the classifier will set the boundaries for each class in the feature space taking into account only the empirical risk aiming at optimizing the separability of the known classes.

\subsubsection{\texorpdfstring{\gls{open}}{Open} strategy}
Depicted in Figure~\ref{fig:open}, in order to let the classifier better tune against unknown samples, a straightforward strategy consists in training the classifier on known data, and tuning it considering both the presence of known and known-unknown samples.
When the open strategy is selected, $\frac{n}{2}$ of the classes are employed as known and the other $\frac{n}{2}$ are employed as known-unknown in validation.
The classifier fitting procedure is carried out on the $\frac{n}{2}$ known classes, however, validation during parameter search is carried out on all $n$ classes, i.e., known and known-unknown camera models.
In doing that, parameter search is performed simulating an open-set setup.
After the best parameters are obtained, the final model is trained with all $n$ known training classes to provide a fair comparison with the \gls{closed} strategy, i.e., the same number of classes to correctly detect is employed.

\subsubsection{\texorpdfstring{\glstext{networkopen}}{NetOpen} strategy}
Depicted in Figure~\ref{fig:netopen}, the \gls{networkopen} strategy employs unknown data---from the point of view of the network used for feature extraction---as known-unknown data for validation.
Dealing with data-driven features (i.e., those extracted using a \gls{cnn}), special attention must be given to the fact that the \gls{cnn}, as a feature extractor, must also be trained and validated on the known classes in order to enable discrimination within the set of known camera models.

This strategy considers that the \gls{cnn} has been separately trained using all available $n$ known classes.
The validation set employed during the \gls{cnn} training process comes from the set of $n$ known classes---as it also happens with \gls{open} and \gls{closed} strategies.
For \gls{networkopen}, to better guide the choice of classifiers' parameters, additionally to the $n$ known classes, the validation set also includes samples from extra known-unknown classes, i.e., classes never employed for \gls{cnn} training or validation.
Parameter search of the classifiers is carried out using all known data along with those extra known-unknown data.
Finally, when hyper-parameters have been selected, the final model training is performed using just the $n$ known classes, for a paired experiment with the other strategies.
In doing that, parameter search is performed simulating an open-set setup also in the point of view of the network.

This approach is appropriate for use with \gls{cnn}-derived features, however, for the sake of fairness, those extra classes, that are known-unknown from the point of view of the network, are also employed in experiments with $\frich$ and $\fcfa$ features when \gls{networkopen} strategy is applied.

\subsection{Open-set Classifiers}
\begin{figure}
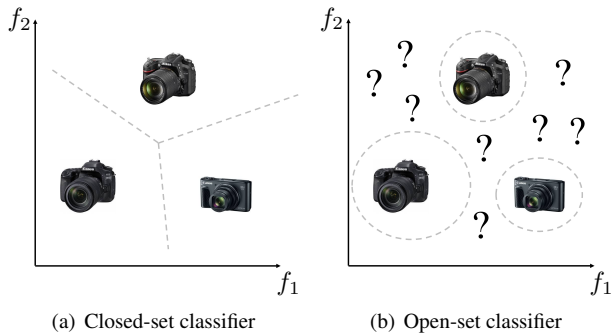

	\centering
	\subfigure[Closed-set classifier]{\includegraphics[width=.45\linewidth]{\changedir{figures/}closed_classifier}\label{fig:closed_classifier}}\hfil
	\subfigure[Open-set classifier]{\includegraphics[width=.45\linewidth]{\changedir{figures/}open_classifier}\label{fig:open_classifier}}
	\caption{%
          Illustration of a three-class classifier in \subref{fig:closed_classifier}~closed- and \subref{fig:open_classifier}~open-set configurations, considering a two-dimensional feature space $[f_1, f_2]$.
          The open-set classifier partitions the feature space with bounded regions referring to the known classes in order to enable rejection of unknown classes.
          In contrast, a closed-set classifier tends to classify any point in the feature space as belonging to one of the known classes.}
	\label{fig:open_closed}
\end{figure}

In the open-set scenario, a classifier should be able to assign one or more \emph{bounded} regions in the feature space for each known class.
In contrast, closed-set classifiers simply splits \emph{unbounded} portions of the feature space to each of the known classes.
This concept is illustrated in Figure~\ref{fig:open_closed}.

In this work, we employ for evaluation multiple open-set classifiers available in the literature.
\Gls{svm} have been applied in the literature to solve various classification problems, including open-set ones in recent works.
Traditional \gls{svm} can be straightforwardly employed for open-set problems by means the one-vs-all \cite{Rocha2014} multiclass-from-binary approach~\cite{MendesJunior2017}: when a feature vector $\f$ is classified as negative by all binary \glspl{svm} that compose the multi-class classifier, then $\f$ is rejected as unknown.
Alternatively, \gls{ocsvm} can also be easily used in open-set setups, as it focuses on carving a decision boundary around known classes, thus points related to unknown classes can be rejected.
The same all-negative criterion can be employed for any one-class classifier~\cite{Heflin2012,Pritsos2013}.
Additionally, other methods derived from \gls{svm} have been proposed in the literature specifically for open-set problems.
In this work, we considered the \gls{wsvm}~\cite{Scheirer2014}, \gls{dbc}~\cite{Costa2012, Costa2014}, \gls{ssvm}~\cite{MendesJunior2018b}, and \gls{pisvm}~\cite{Jain2014}.

In addition to these \gls{svm}-based approaches, we also consider the \Gls{osnn} classifier proposed by \citet{MendesJunior2017}.
This is a recently proposed technique that extends upon the classic nearest neighbors approach.
The main rationale behind this method is to avoid relying on raw similarity scores for thresholding.
Rejection of unknown instances is accomplished through the of ratio of similarity scores instead.
Furthermore, we also consider the classifiers employed by~\citet{Bayar2018}, i.e., \gls{et}~\cite{Geurts2006}, \gls{psvm}, \gls{softmax}, and \gls{ncm}.
Also, by suggestions of previous work~\cite{Bayar2018}, we employ a \gls{2psvm} which consists on having a \gls{ocsvm} for solving the known vs unknown problem, then, if the test instance is classified as known, a \gls{psvm} is employed for choosing the class, otherwise the image is classified to an unknown model.

\section{Experimental Setup}\label{sec:setup}

In this section, we provide details regarding the employed datasets and evaluation metrics.

\subsection{Datasets}
To evaluate all tested methodologies thoroughly, it is important to consider a large enough image database.
In this work, we merged three different datasets freely available from previous work.

\subsubsection{\texorpdfstring{\gls{dresden} \cite{gloe2010dresden}}{Dresden Image Database}}
This dataset contains almost 17000 images from 27 different camera models.
Exactly as in the work of \citet{Bondi2017}, we selected 13000 images from 18 models\footnote{We considered Nikon D70 and Nikon D70s on the same single class due to the negligible differences between them, as reported by \citet{gloe2010dresden}.} as the set of images from known camera models.
This set was split in training, validation, and test sets~\cite{Bondi2017}.
The training set was used to train the \gls{cnn}-based feature extractor and all classifiers.
All images from remaining models---not considered in the subset of 18 models previously selected---have been considered as known-unknown along with the \gls{networkopen} strategy and ignored for both \gls{closed} and \gls{open} strategies.

\subsubsection{\texorpdfstring{\gls{isa}\footnotemark}{Image Source Attribution Unicamp}}
This dataset contains around 9000 images from 35 camera models.
All images from models not overlapping with Dresden Image Database have been selected as unknown-unknown models for the test set in the open-set experiments.
\footnotetext{\label{fn:filipe_dataset}Available at: \url{http://www.recod.ic.unicamp.br/~filipe/dataset}~\cite{Costa2014}.}

\subsubsection{\texorpdfstring{\gls{flickr}\footnotemark[\getrefnumber{fn:filipe_dataset}]}{Flickr Unicamp}}
This dataset comprises around 11000 images from more than 250 camera models.
Differently from previously mentioned datasets, these images have been downloaded from Flickr\footnote{Available at: \url{https://www.flickr.com}.} image hosting service.
To avoid dealing with images from the same camera taken at different resolutions, only images at maximum resolution for each model have been selected.
All images have been considered as belonging to unknown-unknown camera models in test set for the open-set experiments.
\vspace{.5em}

As performed by \citet{Bondi2017}, we obtain, in a content-aware way, 32 non-overlapping $64 \times 64$-pixel patches from each image.
Provided results are based on majority voting after classification per patch.
All patches coming from the same image have been carefully placed only into one of training, validation, and test sets in order to avoid overfitting problems and training/testing contamination.

\subsection{Metrics}

As evaluation metrics, we employ a set of commonly used ones, as well as others recently proposed for open-set scenario \cite{MendesJunior2017,Bayar2018}.
In particular, we consider different definitions of accuracy and f-measure.
Concerning accuracy, we employ the following definitions:

\subsubsection{\texorpdfstring{\gls{aks} \cite{MendesJunior2017}}{AKS}}
This is the accuracy in correctly attributing images from known models to the actual models.
This metric encompasses two kinds of misclassification errors: known-model images attributed to unknown class (false unknown) and known-model images attributed to wrong known classes (misclassification).
\subsubsection{\texorpdfstring{\gls{aus} \cite{MendesJunior2017}}{AUS}}
This is the accuracy in correctly classifying as unknown the images from unknown camera models.
\subsubsection{\texorpdfstring{\gls{na} \cite{MendesJunior2017}}{NA}}
This is the average between \gls{aks} and \gls{aus} and provides an overall view of a classifier performance in terms of both open- and closed-set scenarios.
\subsubsection{\texorpdfstring{\gls{da} \cite{Bayar2018}}{DA}}
This averages the percentage of images from known cameras detected as coming from known models, and the percentage of images from unknown cameras detected as coming from unknown models.
This metric does not take into account whether images from known cameras are misclassified to the wrong camera model.
\vspace{.5em}

Concerning f-measure, an additional comment is in order.
Traditionally, f-measure is defined in terms of precision and recall as
\begin{equation*}
	\textrm{f-measure} = 2 \times \frac{\textrm{precision} \times \textrm{recall}}{\textrm{precision} + \textrm{recall}}.
\end{equation*}
Depending on the definitions of precision and recall employed, we obtain different f-measure definitions.
\citet{MendesJunior2017} has pointed out that it might be inappropriate to consider the \emph{unknown classes} $\co$ as any other known class in terms of \gls{tp}, \gls{fp}, and \gls{fn} calculations.
Therefore, considering $n$ the number of known camera models, and the $(n+1)$-th class concerning the unknown classes $\co$, we resort to the following f-measure definitions:

\subsubsection{\texorpdfstring{\Gls{osfmM} \cite{MendesJunior2017}}{OSFMM}}
F-measure using precision and recall defined as
\begin{equation}
  \label{eq:osfmM}
  \begin{split}
    \textrm{precision} =& \frac{\displaystyle\sum_{i=1}^{n} \frac{\glssymbol{tp}_i}{\glssymbol{tp}_i+\glssymbol{fp}_i} }{n}, \\
    \textrm{recall} =& \frac{\displaystyle\sum_{i=1}^{n} \frac{\glssymbol{tp}_i}{\glssymbol{tp}_i+\glssymbol{fn}_i} }{n}.
  \end{split}
\end{equation}

\subsubsection{\texorpdfstring{\Gls{osfmm} \cite{MendesJunior2017}}{OSFMm}}
F-measure using precision and recall defined as
\begin{equation}
  \label{eq:osfmm}
  \begin{split}
    \textrm{precision} =& \frac{\sum_{i=1}^{n} \glssymbol{tp}_i}{\sum_{i=1}^{n}(\glssymbol{tp}_i+\glssymbol{fp}_i)}, \\
    \textrm{recall} =& \frac{\sum_{i=1}^{n} \glssymbol{tp}_i}{\sum_{i=1}^{n}(\glssymbol{tp}_i+\glssymbol{fn}_i)}.
  \end{split}
\end{equation}

\subsubsection{\texorpdfstring{\Gls{fmM} \cite{Sokolova2009}}{FMM}}
F-measure using precision and recall defined as
\begin{equation*}
  \begin{split}
    \textrm{precision} =& \frac{\displaystyle\sum_{i=1}^{n+1} \frac{\glssymbol{tp}_i}{\glssymbol{tp}_i+\glssymbol{fp}_i} }{n+1}, \\
    \textrm{recall} =& \frac{\displaystyle\sum_{i=1}^{n+1} \frac{\glssymbol{tp}_i}{\glssymbol{tp}_i+\glssymbol{fn}_i} }{n+1}.
  \end{split}
\end{equation*}

\subsubsection{\texorpdfstring{\Gls{fmm} \cite{Sokolova2009}}{FMm}}
F-measure using precision and recall defined as
\begin{equation*}
  \begin{split}
    \textrm{precision} =& \frac{\sum_{i=1}^{n+1} \glssymbol{tp}_i}{\sum_{i=1}^{n+1}(\glssymbol{tp}_i+\glssymbol{fp}_i)}, \\
    \textrm{recall} =& \frac{\sum_{i=1}^{n+1} \glssymbol{tp}_i}{\sum_{i=1}^{n+1}(\glssymbol{tp}_i+\glssymbol{fn}_i)}.
  \end{split}
\end{equation*}
\vspace{.5em}

The main difference between traditional and open-set versions of f-measure is that the latter does not consider the effect of the unknown class in terms of \gls{tp} as the unknown cannot represent a single positive class.
Indeed, the sum index spans the range $[1, n]$ rather than $[1, n+1]$, thus excluding the label $\co$ representing the unknown classes.
However, both \gls{osfmM} and \gls{osfmm} account for \emph{false known} and \emph{false unknown} through \gls{fp} and \gls{fn}, respectively, in Equations~\eqref{eq:osfmM} and \eqref{eq:osfmm}.

\section{Results}\label{sec:results}

We have evaluated all combinations of extracted features (i.e., 5), training protocols (i.e., 3), and classifiers (i.e., 12) for a total amount of 180 cases of study.
Results for each metric are reported in a complete and detailed table of all our experiments, as a supplementary material.\footnote{\label{fn:code}See supplementary material available at \url{https://pedrormjunior.github.io/oscmi.html}.}

Results show that, overall, better performance are obtained for \gls{pisvm}, \gls{et}, and \gls{ssvm} classifiers.
Regarding the training protocols, interestingly, \gls{open} has presented slightly superior results compared to \gls{networkopen}, despite using less known-unknown data.
And, finally, \gls{ip1} presents the better result among the features, although \gls{ip2}, in general, seems to be the most discriminative one.\footnotemark[\getrefnumber{fn:code}]
Hereinafter we report a subset of the obtained results in order to highlight the most interesting findings in terms of best feature set, training protocol, and classifier.

\subsection{Feature Extractors}
To identify the feature vector most suitable for open-set camera model identification problem, we analyzed the behavior of all features (i.e., $\frich$, $\fcfa$, $\fconv$, $\fipone$, and $\fiptwo$) paired with different training strategies and classifiers.
To summarize the achieved results, we rely on \gls{na} as preferred analysis metric.
As a matter of fact, \gls{na} clearly takes into account the ability of correctly classifying known samples at camera level as well as rejecting the unknown.
Therefore, an algorithm with high \gls{na} value is a good candidate to work for both known and unknown classes.

Table~\ref{tab:features} reports the best \gls{na} achieved with each feature extractor.
Specifically, it shows which combination of classifier and training strategy enables to obtain the achieved \gls{na}, as well as all the other metric values for the selected classifier.
From this table, it is possible to notice that the best results are obtained by \gls{cnn}-based features.
In particular, \gls{ip1} achieves the best \gls{na}, which is close to 0.83.
This confirms the behavior observed by \citet{Bondi2017} for the closed-set scenario:
hand-crafted features (i.e., $\frich$ and $\fcfa$) performs better on high resolution images, whereas the \gls{cnn} is superior when trained on small $64 \times 64$ pixel patches as the ones considered in this work.
The explanation for the affected accuracy with hand-crafted features when working with small patches is that hand-crafted features relies on co-occurrences~\cite{Fridrich2012,Chen2015a}, whose computation for small patches might be less stable and reliable.

\begin{table*}[t]
	\centering
	\caption{Best results in terms of \gls{na} achieved with each feature extractor.
          For each metric, the highest results are reported in bold and the lowest ones are reported in italics.}
	\label{tab:features}
	\begin{tabular}{l|ll|l|ll|lllll}
          \hline
          \textbf{Feature} & \textbf{Classifier} & \textbf{Training Protocol} & \textbf{Best \gls{na}}    & \textbf{\gls{aks}}        & \textbf{\gls{aus}}        & \textbf{\gls{da}}         & \textbf{\gls{osfmM}}       & \textbf{\gls{osfmm}}      & \textbf{\gls{fmM}}        & \textbf{\gls{fmm}}        \\
          \hline
          \rowcolor[HTML]{C0C0C0}
          \gls{ip1}        & \gls{pisvm}         & \gls{open}                 & \mynummax{0.827047480908} & \mynummax{0.86388384755}  & \mynum{0.790211114267}    & \mynum{0.811082745157}    & \mynum{0.691604562034}     & \mynummax{0.640991112308} & \mynum{0.70424973862}     & \mynum{0.805451268959}    \\
          \gls{ip2}        & \gls{et}            & \gls{open}                 & \mynum{0.818889480505}    & \mynum{0.858076225045}    & \mynummin{0.779702735965} & \mynummin{0.799819792762} & \mynummax{0.7288136565}    & \mynum{0.630568151507}    & \mynummax{0.737745206605} & \mynummin{0.795915302598} \\
          \gls{conv}       & \gls{pisvm}         & \gls{networkopen}          & \mynum{0.777895619971}    & \mynum{0.687477313975}    & \mynum{0.868313925968}    & \mynummax{0.834659858838} & \mynum{0.663267497466}     & \mynum{0.622003284072}    & \mynum{0.675372055227}    & \mynummax{0.830905541373} \\
          \gls{cfa}        & \gls{ssvm}          & \gls{networkopen}          & \mynum{0.682547923933}    & \mynum{0.46497277677}     & \mynum{0.900123071097}    & \mynum{0.811157831506}    & \mynum{0.356466817652}     & \mynum{0.501860920666}    & \mynum{0.384892460516}    & \mynum{0.810106622616}    \\
          \gls{rich}       & \gls{svm}           & \gls{open}                 & \mynummin{0.576905003305} & \mynummin{0.166969147005} & \mynummax{0.986840859604} & \mynum{0.817465084848}    & \mynummin{0.0844250889815} & \mynummin{0.274054215073} & \mynummin{0.131155212405} & \mynum{0.8172398258}      \\
          \hline
	\end{tabular}
\end{table*}

\begin{table*}[t]
	\centering
	\caption{Best results in terms of \gls{na} achieved with each training protocol.
          For each metric, the highest results are reported in bold and the lowest ones are reported in italics.}
	\label{tab:protocols}
	\begin{tabular}{l|ll|l|ll|lllll}
          \hline
          \textbf{Training Protocol} & \textbf{Feature} & \textbf{Classifier} & \textbf{Best \gls{na}}    & \textbf{\gls{aks}}        & \textbf{\gls{aus}}        & \textbf{\gls{da}}         & \textbf{\gls{osfmM}}      & \textbf{\gls{osfmm}}      & \textbf{\gls{fmM}}        & \textbf{\gls{fmm}}        \\
          \hline
          \rowcolor[HTML]{C0C0C0}
          \gls{open}                 & \gls{ip1}        & \gls{pisvm}         & \mynummax{0.827047480908} & \mynum{0.86388384755}     & \mynum{0.790211114267}    & \mynum{0.811082745157}    & \mynummax{0.691604562034} & \mynummax{0.640991112308} & \mynummax{0.70424973862}  & \mynum{0.805451268959}    \\
          \gls{closed}               & \gls{ip1}        & \gls{ssvm}          & \mynum{0.799333821632}    & \mynummax{0.873684210526} & \mynummin{0.724983432737} & \mynummin{0.762276618111} & \mynummin{0.627453808685} & \mynummin{0.590385087074} & \mynum{0.645340908518}    & \mynummin{0.755744105722} \\
          \gls{networkopen}          & \gls{ip2}        & \gls{softmax}       & \mynummin{0.784707363803} & \mynummin{0.720508166969} & \mynummax{0.848906560636} & \mynummax{0.822420783901} & \mynum{0.627553896222}    & \mynum{0.626479406659}    & \mynummin{0.641033160905} & \mynummax{0.822345697552} \\
          \hline
	\end{tabular}
\end{table*}

\begin{table*}[t]
	\centering
	\caption{Best results in terms of \gls{na} achieved with each open-set classifier.
          For each metric, the highest results are reported in bold and the lowest ones are reported in italics.}
	\label{tab:classifiers}
	\begin{tabular}{l|ll|l|ll|lllll}
          \hline
          \textbf{Classifier} & \textbf{Feature} & \textbf{Training Protocol} & \textbf{Best \gls{na}}    & \textbf{\gls{aks}}         & \textbf{\gls{aus}}        & \textbf{\gls{da}}         & \textbf{\gls{osfmM}}       & \textbf{\gls{osfmm}}       & \textbf{\gls{fmM}}        & \textbf{\gls{fmm}}        \\\hline
          \rowcolor[HTML]{C0C0C0}
          \gls{pisvm}         & \gls{ip1}        & \gls{open}                 & \mynummax{0.827047480908} & \mynum{0.86388384755}      & \mynum{0.790211114267}    & \mynum{0.811082745157}    & \mynum{0.691604562034}     & \mynum{0.640991112308}     & \mynum{0.70424973862}     & \mynum{0.805451268959}    \\
          \gls{et}            & \gls{ip2}        & \gls{open}                 & \mynum{0.818889480505}    & \mynum{0.858076225045}     & \mynum{0.779702735965}    & \mynum{0.799819792762}    & \mynummax{0.7288136565}    & \mynum{0.630568151507}     & \mynummax{0.737745206605} & \mynum{0.795915302598}    \\
          \gls{ssvm}          & \gls{ip1}        & \gls{closed}               & \mynum{0.799333821632}    & \mynummax{0.873684210526}  & \mynum{0.724983432737}    & \mynum{0.762276618111}    & \mynum{0.627453808685}     & \mynum{0.590385087074}     & \mynum{0.645340908518}    & \mynum{0.755744105722}    \\
          \gls{softmax}       & \gls{ip2}        & \gls{networkopen}          & \mynum{0.784707363803}    & \mynum{0.720508166969}     & \mynum{0.848906560636}    & \mynum{0.822420783901}    & \mynum{0.627553896222}     & \mynum{0.626479406659}     & \mynum{0.641033160905}    & \mynum{0.822345697552}    \\
          \gls{osnn}          & \gls{ip2}        & \gls{networkopen}          & \mynum{0.784087987845}    & \mynum{0.681306715064}     & \mynum{0.886869260627}    & \mynum{0.844421084247}    & \mynum{0.634360528652}     & \mynummax{0.644131777625}  & \mynum{0.648523867855}    & \mynummax{0.844345997898} \\
          \gls{svm}           & \gls{ip1}        & \gls{open}                 & \mynum{0.762558363414}    & \mynum{0.757531760436}     & \mynum{0.767584966392}    & \mynum{0.767607748911}    & \mynum{0.530522701813}     & \mynum{0.569829351536}     & \mynum{0.551414413744}    & \mynum{0.765505331131}    \\
          \gls{psvm}          & \gls{conv}       & \gls{open}                 & \mynum{0.75443604899}     & \mynum{0.715063520871}     & \mynum{0.793808577109}    & \mynum{0.779246133053}    & \mynum{0.603295915816}     & \mynum{0.568870921167}     & \mynum{0.617996221779}    & \mynum{0.777519147019}    \\
          \gls{ncm}           & \gls{ip2}        & \gls{open}                 & \mynum{0.733938345555}    & \mynum{0.852994555354}     & \mynum{0.614882135757}    & \mynum{0.672248085298}    & \mynum{0.637937586676}     & \mynum{0.506410947096}     & \mynum{0.651430279966}    & \mynum{0.664138759574}    \\
          \gls{ocsvm}         & \gls{ip2}        & \gls{open}                 & \mynum{0.674193212516}    & \mynum{0.387295825771}     & \mynum{0.961090599262}    & \mynummax{0.847124192822} & \mynum{0.504835318545}     & \mynum{0.496741154562}     & \mynum{0.533305285499}    & \mynum{0.842393752816}    \\
          \gls{dbc}           & \gls{conv}       & \gls{open}                 & \mynum{0.637123864711}    & \mynum{0.790199637024}     & \mynummin{0.484048092398} & \mynummin{0.553536567052} & \mynum{0.512768076137}     & \mynum{0.416093272171}     & \mynum{0.533469940303}    & \mynummin{0.547379486409} \\
          \gls{2psvm}         & \gls{cfa}        & \gls{closed}               & \mynum{0.588058924991}    & \mynum{0.384392014519}     & \mynum{0.791725835463}    & \mynum{0.713019972969}    & \mynum{0.361535929612}     & \mynum{0.347897503285}     & \mynum{0.385358137818}    & \mynum{0.707463583121}    \\
          \gls{wsvm}          & \gls{cfa}        & \gls{open}                 & \mynummin{0.507852667935} & \mynummin{0.0286751361162} & \mynummax{0.987030199754} & \mynum{0.795314611804}    & \mynummin{0.0748332561292} & \mynummin{0.0516677567037} & \mynummin{0.129464815291} & \mynum{0.788782099414}    \\
          \hline
	\end{tabular}
\end{table*}

It is interesting to notice how \gls{aks} and \gls{aus} are unbalanced for hand-crafted features.
For instance, \gls{rich} and \gls{cfa} show \gls{aus} higher than 0.90, but \gls{aks} lower than 0.50.
This means that the classifier rejects many more images as unknown than it should.
This makes these features not appealing for open-set problems, as the presence of unknown devices greatly hinders the closed-set classification capability of these features.
The same behavior is also captured by the metrics based on f-measure.
Conversely, \gls{ip1} is able to correctly classify unknown images with almost 0.80 accuracy (\gls{aus}), and to correctly attribute known-camera images to their model with 0.86 accuracy (\gls{aks}).

\subsection{Training Protocols}
To evaluate the different training protocols, we considered \gls{na} as reference metric for the same reasons previously mentioned.
Table~\ref{tab:protocols} reports the best \gls{na} results for each protocol, also showing which feature and classifier is used to obtain the reported result.
Also, the other metrics are then reported for each case.

It is possible to notice that \gls{open} strategy presents better results, more than 4\% higher than the best result with \gls{networkopen}.
In Table~\ref{tab:protocols}, although \gls{closed} strategy presents better results than \gls{networkopen}, in general, we have observed that \gls{closed} tends to perform the worse.\footnotemark[\getrefnumber{fn:code}]
Also in a general evaluation, we also observe that, in fact, \gls{open} tends to perform slightly better than \gls{networkopen}.

It is worth to highlight one aspect about the \gls{closed} strategy.
Despite this strategy's name, all classifiers employed along with it are open-set ones.
Therefore, even if trained only considering known camera images, they still have the ability to reject new data as unknown (remember, from Section~\ref{sec:training-protocol}, the different training protocols refers only to the split of the training data).
This explains why using the \gls{closed} strategy is still possible to achieve \gls{aus} higher than 0.70.
However, even though, \gls{open} approaches 0.80, almost 10\% of difference from the \gls{closed} strategy is observed.

Furthermore, considering all the 360 measured combinations ($12 \text{ classifiers} \times 5 \text{ features sets} \times 6 \text{ metrics}$), classifiers training with \gls{open} obtained better results than versions trained with \gls{networkopen} in 188 of the them, while \gls{networkopen} wins in 172 cases.
It also indicates a slightly better performance for \gls{open} protocol.
However, when \gls{networkopen} achieves better results, the classifiers obtain an average of about 10.5\% better results, while \gls{open} improves only 7.8\% in average.

This is a counter intuitive result, as \gls{networkopen} uses the same known data as \gls{open} strategy does, along with extra known-unknown data from the other \gls{dresden} classes not employed as known.
The numbers regarding the difference of those two training protocols indicates some similarity among the representativeness of the two sets of training data.
Therefore, those results indicate that by simply having some known-unknown data, although they are not unknown from the point of view of the network (\gls{open} strategy), is enough for improving the performance compared to the traditional \gls{closed} form.
It means those extra data are not necessary, which is a good trace also for making the training process cheaper.

Moreover, those results are good evidences that representation of unknown instances are as distinct as the representations of known-unknown from the point of view of the network.
It means those representations are distinct alike from the known instances after a trained network is employed for feature extraction.
Those results are also in tune with the ones presented by~\citet{Bondi2017}:
they have performed a closed-set experiment with a distinct set of camera models not employed for network training and they have showed that representations for those camera models are distinct enough to allow discrimination among them.

\subsection{Open-set Classifiers}\label{sec:results-open-set-classifiers}
To analyze the effect of different classifiers, Table~\ref{tab:classifiers} reports the best \gls{na} result obtained with each classifier, showing also the feature and training strategy used in each case.
For each selected methodology, all other metrics are also reported.

From these results---as we saw in other tables as well---it is possible to see that \gls{pisvm} performs better than its counterparts, achieving \gls{na} close to 0.83, however, best \gls{aks} and \gls{aus} are obtained with \gls{ssvm} and \gls{ocsvm}, respectively.
Results in Table~\ref{tab:classifiers} show many classifiers with reasonable performance: among the cases, \gls{et} have obtained the best performance for the macro-averaging versions of the f-measure measures and \gls{osnn} presents best results for the micro-averaging versions.
\gls{ocsvm} also outperforms other methods based on \gls{da} although its high propensity of rejecting instances as unknown.
Additionally, \gls{closed} protocol only appears to be the best one for \gls{ssvm} and \gls{2psvm} classifiers, all other classifiers has the \gls{open} or \gls{networkopen} variations as the best training protocol, and \gls{open} appears in most of the cases.

It is important to notice that \gls{2psvm} appears as one of the last methods in the ranking of Table~\ref{tab:classifiers}.
This low performance for \gls{2psvm} can be justified by its implicit assumption that all known classes can be modeled as a single class.
It does not take into account the fact that known classes can be sparse in the feature space and some intermediate regions among those classes can refer to the unknown, i.e., it is difficult to specialize on the known classes by means of a single model.
Furthermore, the best \gls{na} result with \gls{2psvm} is obtained with \gls{closed} training protocol, which indicates that even though simulation of the open-set scenario is performed for parameter optimization, a one-class classifier is not able to handle well the feature space.

In general, we verify that by the straightforward employment of an open-set classifier, as is, improves results for the open-set scenario compared to closed-set classifiers adapted for open-set recognition by means rejection through thresholding of similarity scores.
Further details regarding comparison with those state-of-the-art solutions are presented in the next section.

\subsection{Comparison with State-of-the-art}\label{sec:sota}
\begin{table*}[t]
	\centering
	\caption{Difference achieved by the best solution found through the pipeline considered in this work (\gls{pisvm}) and the baselines.
          Results obtained for \gls{ip1} feature for the corresponding methods implemented along with \gls{open} training protocol.
          The consistency of positive values for $\Delta$ evinces the improvement of the found solution over the state-of-the-art methods.}
	\label{tab:sota}
	\begin{tabular}{l|llllllll}
          \hline
          \textbf{Reference} & \textbf{$\Delta$\gls{na}} & \textbf{$\Delta$\gls{aks}} & \textbf{$\Delta$\gls{aus}}  & \textbf{$\Delta$\gls{da}}   & \textbf{$\Delta$\gls{osfmM}} & \textbf{$\Delta$\gls{osfmm}} & \textbf{$\Delta$\gls{fmM}} & \textbf{$\Delta$\gls{fmm}}   \\
          \hline
          \gls{softmax}      & \mynum{0.327047480908}    & \mynum{0.86388384755}      & \mynummin{-0.209788885733}  & \mynum{0.0179456374831}     & \mynum{0.691604562034}       & \mynum{0.640991112308}       & \mynum{0.657689932211}     & \mynum{0.0123141612855}      \\
          \gls{et}           & \mynum{0.0547203512999}   & \mynum{0.164065335753}     & \mynummin{-0.0546246331535} & \mynummin{-0.0044300946088} & \mynum{0.0203860633455}      & \mynum{0.0319246141246}      & \mynum{0.0220796233242}    & \mynummin{-0.00938579366271} \\
          \gls{ncm}          & \mynum{0.169830004503}    & \mynum{0.0747731397459}    & \mynum{0.264886869261}      & \mynum{0.220528607899}      & \mynum{0.137467614531}       & \mynum{0.209855762779}       & \mynum{0.132279552415}     & \mynum{0.225559393302}       \\
          \gls{psvm}         & \mynum{0.09376728996}     & \mynum{0.120508166969}     & \mynum{0.0670264129509}     & \mynum{0.0770385943835}     & \mynum{0.0512896168352}      & \mynum{0.116937888153}       & \mynum{0.0512055137652}    & \mynum{0.0780898032738}      \\
          \hline
	\end{tabular}
\end{table*}
To the best of our knowledge, the only work presenting results for the open-set camera model identification problem is the work of \citet{Bayar2018}.
In particular, in this work, the authors propose two different approaches.
The first one (\ref{sec:approach1}) relies on \emph{confidence score thresholding}: when the classifier is not ``sure'' about its classification to a certain known class, test instance is then rejected as unknown.
The second approach (\ref{sec:approach2}) assumes known-unknown data is available for training a classifier for detecting if a test instance is known or unknown.
For this approach, previous work have evaluated only the detection ability although in a real open-set scenario further decision should be required to chose the correct class in case an instance is detected as known.

\subsubsection{Approach 1}\label{sec:approach1}
The first approach proposed by \citet{Bayar2018} works as it follows.
A multi-class classifier is trained with \gls{closed} training protocol.\footnote{Previous work \cite{Bayar2018} have not evaluated neither \gls{open} nor \gls{networkopen} protocols.  To the best of our knowledge, our work evaluates them for the first time in this problem.}
This classifier is chosen in order to also provide a confidence score about detected class.
Instances providing a low confidence score are classified as unknown.
For this class of methods, we implemented their solutions based on \glsfirst{softmax}, \glsfirst{ncm}, \glsfirst{psvm}, and \gls{et}~\cite{Geurts2006}.

Table~\ref{tab:sota} reports the metric difference $\Delta$ achieved by the best solution we have evaluated in previous sections compared to the baselines, by considering, for each method, the setup that maximizes \gls{na}.
From this comparison, in general, it is possible to notice that the best solution we found in our analysis is able to achieve better results than all strategies reported by \citet{Bayar2018}.
For most of the measures, for each of the compared baselines, \gls{pisvm} improves the accuracy.

\begin{figure}
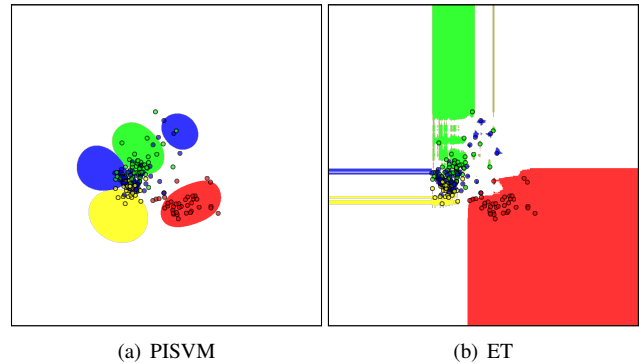

  \centering
  \subfigure[\gls{pisvm}]{\includegraphics[frame,width=0.48\linewidth]{\changedir{boundaries/}figBoundaries__camera_networkopen_ip2_forboundaries__pisvm_ovx_gseo__color__800__1_0}\label{fig:boundaries-pisvm}}
  \subfigure[\gls{et}]{\includegraphics[frame,width=0.48\linewidth]{\changedir{boundaries/}figBoundaries__camera_networkopen_ip2_forboundaries__mcet_imc_gseo__color__800__1_0}\label{fig:boundaries-et}}
  \caption{Decision boundaries of the \gls{pisvm} open-set classifier compared to competing \gls{et} classifier.}
  \label{fig:boundaries}
\end{figure}

We see in the same table that \gls{et}, as employed by \citet{Bayar2018}, is the most competitive method compared to a classifier specially designed for open-set scenario (\gls{pisvm}).
Although its high accuracy, we should analyze some theoretical properties of the classifier.
For instance, consider the ability of bounding the region of the feature space in which a possible test instance would be classified as belonging to one of the known classes, i.e., bounding the \gls{klos}~\cite{Scheirer2013,MendesJunior2017}.
Figures~\ref{fig:boundaries-pisvm} and \ref{fig:boundaries-et} depict the decision boundaries of \gls{pisvm} and \gls{et} classifiers, respectively, in the feature space formed by the two first features of the \gls{ip2} layer.
For those images, only training samples from the 4 first classes, out of the 18, were employed to avoid cluttering the visualization.
Small circles represent training samples.
Colored regions indicate that a possible test instance in there would be classified to the class of the same color.
The white region represents rejection as unknown.
In Figure~\ref{fig:boundaries}, we observe that \gls{pisvm} is able to bound the \gls{klos}, properly ensuring the rejection of any data point that would appear far from the support of the training samples in the feature space.
However, by thresholding the probability score of the \gls{et} classifier, the same property is not ensured.
In general, we see that \gls{pisvm} demonstrates a more controlled behavior.

\subsubsection{Approach 2}\label{sec:approach2}
The second approach proposed by \citet{Bayar2018} works as it follows.
A binary classifier is trained to distinguish between images from known and unknown camera models.
The objective here is to analyze only the detection ability.
All samples from all known classes are considered into a single known class called \emph{known}.
Extra data from other classes not of interest are employed on the \emph{unknown} class for the binary classification.
As in the previous experiments, we consider the 18 classes of \gls{dresden} as the known classes of interest.
For the extra known-unknown data, we employed the remaining classes of the \gls{dresden} dataset, as those classes were also employed along with \gls{networkopen} training protocol.
For this method, we implemented both solutions shown by~\citet{Bayar2018}, i.e., \gls{psvm}\footnote{Notice, however, that Platt's probability is not required to be employed in this context as only the class decision matters in this case.} and \gls{et}.

In Table~\ref{tab:sota-detection}, we present the \gls{da} for the two baselines as well as for \gls{pisvm} solution which have presented the best results throughout our experiments.
Furthermore, \gls{dks} and \gls{dus} are also presented for a more in-depth evaluation of the performance of the classifiers.
\gls{networkopen} was selected for those results because baselines require extra known-unknown data for training although \gls{pisvm} have obtained best performance along with \gls{open} strategy (Section~\ref{sec:results-open-set-classifiers}).
\gls{ip2} is employed in this case because, as previously saw\footnotemark[\getrefnumber{fn:code}], it has comparable or better results than \gls{ip1} in general and, furthermore, baselines has presented slightly better results with this feature, compared to \gls{ip1}.

Our results for this approach, as seen in Table~\ref{tab:sota-detection}, are far from the ones reported by \citet{Bayar2018} as the baselines have \emph{almost} no ability to reject instances as unknown.
Our conclusions from those results is that by relying solely on known-unknown data for training a classifier to distinguish, in the wild, known versus unknown classes is susceptible to a worst case scenario.
We conjecture that the known-unknown data employed for those classifiers makes them create a decision frontier in the feature space in such a way that most of the real unknown data (from \gls{isa} and \gls{flickr} datasets) becomes accepted as known.
If a different set of known-unknown is employed in place of the unknown part of the \gls{dresden} dataset, we believe results might drastically differ.
Taking an essentially distinct approach, \gls{pisvm} along with \gls{networkopen} training protocol does not rely solely on the known-unknown data for defining its boundary decision: instead, it minimizes the \emph{risk of the unknown} also taking advantage of the inter-class information gathered from the known data~\cite{Jain2014}.

\begin{table}[t]
  \centering
  \caption{Difference achieved by the best solution found through the pipeline considered in this work (\gls{pisvm}) and the baselines considering Approach 2 of~\citet{Bayar2018}.
    Results obtained for \gls{ip2} feature for the corresponding methods implemented along with \gls{networkopen} training protocol.
    For each metric, the highest results are reported in bold.}
  \label{tab:sota-detection}
  \begin{tabular}{l|lll}
    \hline
    \textbf{Reference} & \textbf{\glstext{da}}     & \textbf{\glstext{dks}} & \textbf{\glstext{dus}}    \\
    \hline
    \gls{pisvm}        & \mynummax{0.741853131101} & \mynum{0.88058076225}  & \mynummax{0.70567073748}  \\
    \gls{psvm}         & \mynum{0.206862892326}    & \mynummax{1.0}         & \mynum{0.0}               \\
    \gls{et}           & \mynum{0.208740051059}    & \mynummax{1.0}         & \mynum{0.00236675186973}  \\
    \hline
  \end{tabular}
\end{table}

\subsection{Post-fusion Analysis}
In the machine learning field, it is well known that jointly using a series of different models can help increasing classification performance.
This is known as ensemble learning~\cite{Sagi2018}.
In the light of this, here we present results achieved with a very simple yet effective ensemble fusion technique.
We perform majority voting among different models.
Given a set of trained models, we test the image under analysis with all of them, and perform majority voting on their output.
If the majority of the votes is for rejecting as unknown, the image is then classified as unknown.

By considering all 1048575 combinations obtained by fusing up to 8 single models achieving \gls{na} greater than 0.7.
Top-three results are reported in Table~\ref{tab:post_fusion}.
Notice that, the features that are selected are always $\fipone$ and $\fiptwo$.
Moreover, top results includes all three training protocols.
The classifiers that appear among those selected solutions are \gls{pisvm}, \gls{ssvm}, \gls{osnn}, and \gls{et}.
These results confirm that by using post-fusion it is actually possible to increase \gls{na} of approximately 2.5\%, and no more than 6 models are needed.
This paves the way to the development of more complex ensemble methods for open-set camera model identification.

\begin{table*}[]
	\centering
	\caption{Top three post-fusion results in terms of \gls{na} by considering the alternatives with accuracy greater than 0.7 and the fusion of at most 8 models.}
	\label{tab:post_fusion}
        \resizebox{\linewidth}{!}{
          \begin{tabular}{l|l|l}
            \hline
            \textbf{Combination}                                                                                                                                                                                                                  & \textbf{N. models} & \textbf{\gls{na}} \\ \hline
            (\gls{open}, \gls{osnn}, \gls{ip2}), (\gls{networkopen}, \gls{pisvm}, \gls{ip2}), (\gls{open}, \gls{ssvm}, \gls{ip2}), (\gls{closed}, \gls{ssvm}, \gls{ip1}), (\gls{open}, \gls{et}, \gls{ip2}), (\gls{open}, \gls{pisvm}, \gls{ip1}) & \mynum{6}          & \mynum{0.85215}   \\
            (\gls{networkopen}, \gls{ssvm}, \gls{ip1}), (\gls{open}, \gls{osnn}, \gls{ip2}), (\gls{open}, \gls{ssvm}, \gls{ip2}), (\gls{closed}, \gls{ssvm}, \gls{ip1}), (\gls{open}, \gls{et}, \gls{ip2}), (\gls{open}, \gls{pisvm}, \gls{ip1})  & \mynum{6}          & \mynum{0.85205}   \\
            (\gls{open}, \gls{osnn}, \gls{ip2}), (\gls{open}, \gls{ssvm}, \gls{ip2}), (\gls{closed}, \gls{ssvm}, \gls{ip1}), (\gls{open}, \gls{et}, \gls{ip2}), (\gls{open}, \gls{pisvm}, \gls{ip1})                                              & \mynum{5}          & \mynum{0.85193}   \\\hline
          \end{tabular}
        }
\end{table*}

\subsection{Impact of an open-set solution}

\begin{figure}[t]
	\centering
	\subfigure[Open-set solution with \gls{pisvm} on \gls{ip1} features along with \gls{open} training protocol.]{\includegraphics[width=\linewidth]{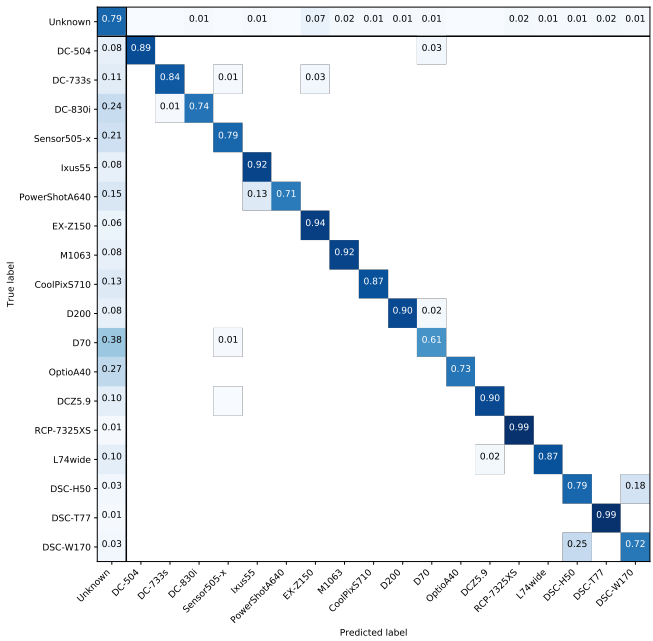}\label{fig:cm-open}}
	\subfigure[Close-set solution as the output of the network, i.e., without reject option.]{\includegraphics[width=\linewidth]{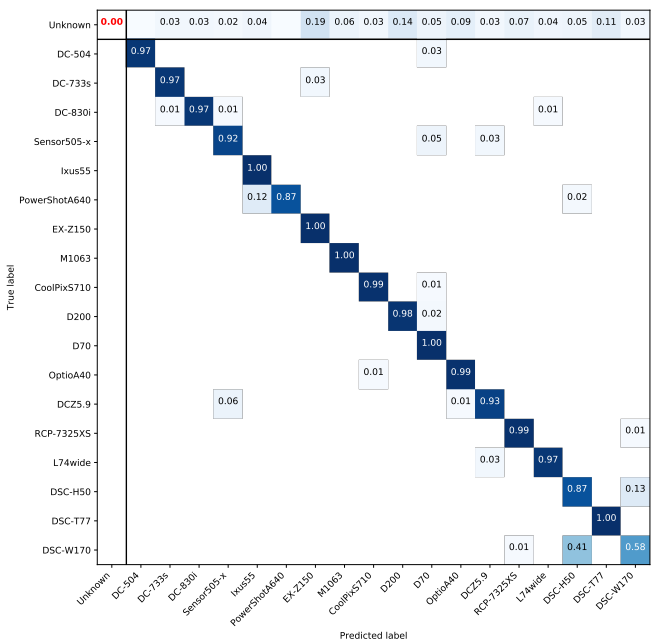}\label{fig:cm-closed}}
	\caption{Confusion matrix comparing the best open-set results against the baseline close-set \gls{cnn} solution~\cite{Bondi2017}, that is equivalent to the \gls{softmax} method~\cite{Bayar2018} without the rejection threshold.}
	\label{fig:cmatrix}
\end{figure}

In Figure~\ref{fig:cmatrix}, we present two confusion matrices.
One of them, in Figure~\ref{fig:cm-open}, obtained by an open-set solution and the other, in Figure~\ref{fig:cm-closed}, by the closed-set output of the neural network employed along this work.
By comparing Figures~\ref{fig:cm-open} and \ref{fig:cm-closed}, we observe that the ability on recognizing instances of each individual model is affected on the open-set solution.
That is expect as long as an open-set solution can also perform the fault of rejecting instances as unknown, i.e., \emph{false unknown}, while a closed-set solution can only have \emph{misclassifications}.
On the other hand, we clearly see the undesirable behavior of the closed-set solution of assigning every unknown instance to one of the known models, i.e., 0\% on \gls{aus} or, in another perspective, 100\% of \emph{false known}.
The false known rate obtained by the open-set solution, in this example, is 21\%.
Anyhow, it is worth noticing that most of the open-set classifiers can be tuned to decrease its false known rate although with the expense of increasing its false unknown rate.

\section{Conclusions}\label{sec:conclusions}

In this paper, we studied the use of a supervised-learning strategy for image camera model identification in an open-set scenario.
In doing so, we explored the possibility of using multiple camera-related features proposed in the literature for closed-set camera model identification, however, under the more challenging open-set regime.
We considered pairing feature vectors with different open-set classifiers exploring also the use of three alternatives of training protocols.
All tests have been performed considering a selection of three independent image datasets freely available online comprising a large number of images from more than 300 camera models.

In terms of training protocols, we found out that employing extra known-unknown classes, as for \gls{networkopen} approach, in general does not help on improving the performance of the classifiers compared to the simpler and cheaper employment of the \gls{open} strategy.
This result is interesting as it evinces that extra known-unknown classes, from the point of view of the network, are not required to be employed as its impact is limited.
It means one can successfully train any open-set classifier, along with an \gls{open} training protocol, with only the data available for the known classes.
A better intuition on this behavior requires a deeper study on the network's representation for unknown classes not employed on network training and those should be compared among the representation of each of the known classes employed for training the network, therefore, it remains as a future work.

Another evidence on the limited use of the known-unknown data from the point of view of the network were presented by employing a binary classifier for recognizing known versus unknown camera models: when a known-unknown set of data (from the unknown part of \gls{dresden}) is employed to train this classifier, its performance on detecting unknown camera models from \gls{isa} and \gls{flickr} datasets is highly effected (Section~\ref{sec:approach2}).
It also reinforces previous arguments on the open-set area that more theoretically-sounded and less data-relied solutions should be developed for general open-set problems~\cite{Scheirer2013}.

Our results have shown that appropriate means of dealing with the open-set camera model attribution problem should be sought in order to properly handling the problem, considering that a recently proposed open-set method~\cite{Jain2014}, as is, obtains considerable improved results compared to the straightforward idea of thresholding the softmax probability of neural networks for rejection as unknown (Section~\ref{sec:approach1}).
This problem on thresholding the softmax probability for open-set recognition have been evinced in one of our previous work, hence the current work also confirms the previously more theoretical perspective~\cite[Chapter 7]{MendesJunior2018b}.

For the open-set camera model identification problem, a promising future research can be performed on investigating recently proposed alternatives to the softmax loss, e.g., the center loss~\cite{Wen2019}, the angular softmax loss~\cite{Liu2017}, etc., as the authors of those works have claimed improvement on the open-set face recognition problem.


\bibliographystyle{IEEEtranN}
\bibliography{biblio}

\begin{IEEEbiography}[{\includegraphics[width=1in,height=1.25in,clip,keepaspectratio]{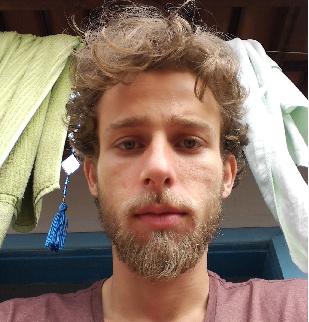}}]{Pedro Ribeiro Mendes Júnior,} Brazilian, received his B.Sc.\@ in Computer Science from Federal University of Ouro Preto (UFOP), Ouro Preto, Minas Gerais, Brazil, in 2012 and his M.Sc.\@ and Ph.D., also in Computer Science, from the University of Campinas (UNICAMP), Campinas, São Paulo, Brazil, in 2014 and 2018, respectively. He is currently a postdoc fellow at the Image and Sound Processing Group, Politecnico di Milano, Milan, Italy, working on the project DeepEyes financed by Coordenação de Aperfeiçoamento de Pessoal de Nível Superior (CAPES) from Brazil. His research interests focus on open-set recognition and, currently, open-set methods and techniques applied to multimedia forensics applications.
\end{IEEEbiography}

\begin{IEEEbiography}[{\includegraphics[width=1in,height=1.25in,clip,keepaspectratio]{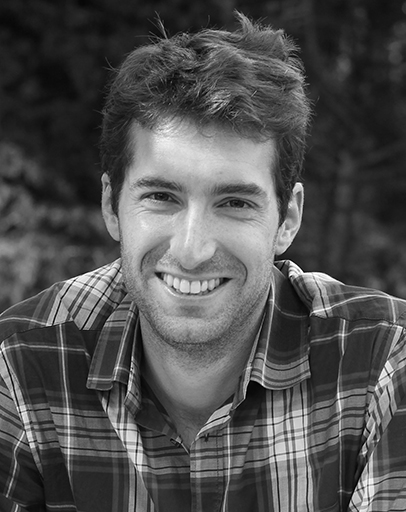}}]{Luca Bondi} received the M.Sc.\@ in Computer Science and Engineering and the Ph.D.\@ in Information Technology from the Politecnico di Milano, Milan, Italy, in 2014 and 2019, respectively. He is currently a postdoc fellow at the Image and Sound Processing Group, Politecnico di Milano. His research interests focus on data-driven methods applied to images and videos in multimedia forensics applications.
\end{IEEEbiography}

\begin{IEEEbiography}[{\includegraphics[width=1in,height=1.25in,clip,keepaspectratio]{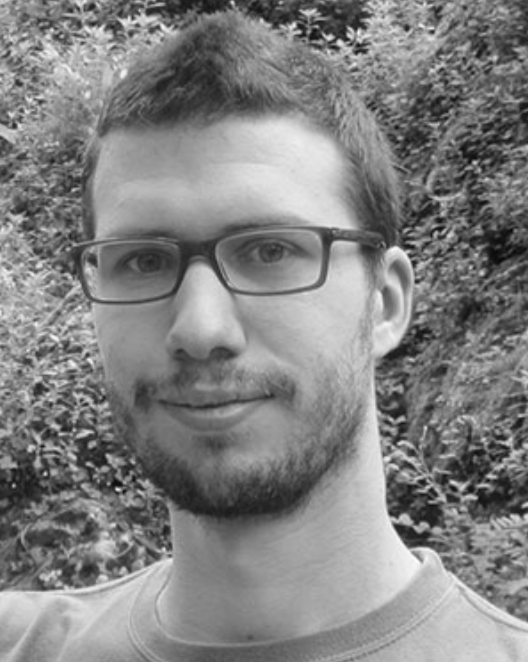}}]{Paolo Bestagini} was born in Novara, Italy, on February 22, 1986. He received the M.Sc.\@ degree in Telecommunications Engineering and the Ph.D.\@ degree in Information Technology from the Politecnico di Milano, Italy, in 2010 and 2014, respectively. He is currently an Assistant Professor at the Image and Sound Processing Group, Politecnico di Milano. His research interests focus on multimedia forensics and acoustic signal processing for microphone arrays. He is an elected member of the IEEE Information Forensics and Security Technical Committee, and a co-organizer of the IEEE Signal Processing Cup 2018.
\end{IEEEbiography}

\begin{IEEEbiography}[{\includegraphics[width=1in,height=1.25in,clip,keepaspectratio]{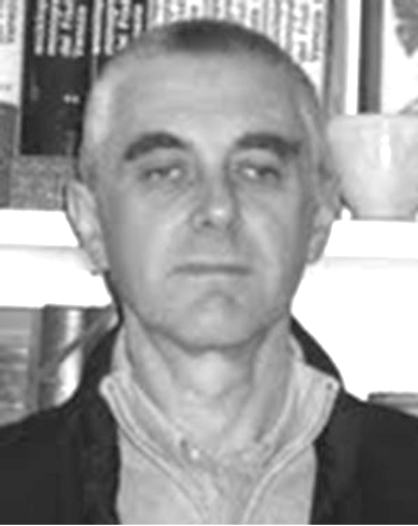}}]{Stefano Tubaro} was born in Novara, Italy, in 1957. He completed his studies in Electronic Engineering at the Politecnico di Milano, Milan, Italy, in 1982. He then joined the Dipartimento di Elettronica, Informazione e Bioingegneria of the Politecnico di Milano, first as a Researcher of the National Research Council, and then (in November 1991) as an Associate Professor. Since December 2004, he has been appointed as a Full Professor of telecommunication at the Politecnico di Milano. His current research interests include advanced algorithms for video and sound processing. He is the author of more than 180 scientific publications on international journals and congresses and the coauthor of more than 15 patents. In the past few years, he has focused his interest on the development of innovative techniques for image and video tampering detection and, in general, for the blind recovery of the “processing history” of multimedia objects. He coordinates the research activities of the Image and Sound Processing Group at the Dipartimento di Elettronica, Informazione e Bioingegneria, Politecnico di Milano. He had the role of Project Coordinator of the European Project ORIGAMI: A new paradigm for high-quality mixing of real and virtual and of the research project ICT-FET-OPEN REWIND: REVerse engineering of audio-VIsual coNtent Data. This last project was aimed at synergistically combining principles of signal processing, machine learning, and information theory to answer relevant questions on the past history of such objects. He is a member of the IEEE Multimedia Signal Processing Technical Committee and of the IEEE SPS Image Video and Multidimensional Signal Technical Committee. He was in the organization committee of a number of international conferences including the IEEE MMSP 2004/2013, IEEE ICIP 2005, IEEE AVSS 2005/2009, IEEE ICDSC 2009, IEEE MMSP 2013, IEEE ICME 2015. From May 2012 to April 2015, he was an Associate Editor of the IEEE Transactions on Image Processing, and is currently an Associate Editor of the IEEE Transactions on Information Forensics and Security.
\end{IEEEbiography}

\begin{IEEEbiography}[{\includegraphics[width=1in,height=1.25in,clip,keepaspectratio]{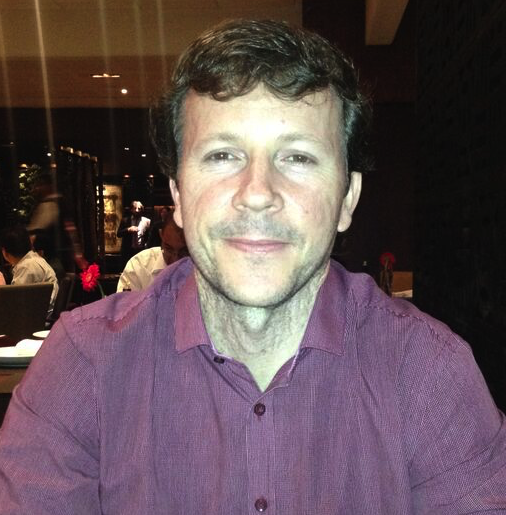}}]{Anderson Rocha} is an associate professor at the Institute of Computing, University of Campinas (UNICAMP), Brazil. His main interests include Digital Forensics, Reasoning for Complex Data and Machine Intelligence. He has actively worked as a program committee member in several important Computer Vision, Pattern Recognition, and Digital Forensics events and is an associate editor of important international journals such as the IEEE Transactions on Information Forensics and Security (T.IFS), Elsevier Journal of Visual Communication and Image Representation (JVCI), the EURASIP/Springer Journal on Image and Video Processing (JIVP) and the IEEE Security \& Privacy Magazine. He is an elected affiliate member of the
Brazilian Academy of Sciences (ABC) and the Brazilian Academy of Forensic Sciences (ABC). He is a two-term elected member of the IEEE Information Forensics and Security Technical Committee (IFS-TC) its chair for 2019/2020 term. He is a Microsoft Research and a Google Research Faculty Fellow, important academic recognitions given to researchers by Microsoft Research and Google, respectively. In addition, in 2016, he has been awarded the Tan Chin Tuan (TCT) Fellowship, a recognition promoted by the Tan Chin Tuan Foundation in Singapore. He has been the principal investigator of a number of research projects in partnership with public funding agencies in Brazil and abroad as well as national and multi-national companies having already deposited and licensed several patents. He is a Brazilian CNPq research scholar (PQ1). Since March 2019, he has been the Director of the Institute of Computing, University of Campinas (UNICAMP).
\end{IEEEbiography}

\EOD

\end{document}